\pdfoutput=1

\documentclass[11pt]{article}

\usepackage{acl}

\usepackage{times}
\usepackage{latexsym}
\usepackage{natbib}
\usepackage{booktabs}
\usepackage{multirow}
\usepackage{multicol}
\usepackage{lipsum}
\usepackage{bbm}
\usepackage{natbib}

\usepackage{hyperref}
\usepackage{url}
\usepackage{graphicx}
\usepackage{times}
\usepackage{latexsym}
\usepackage{natbib}
\usepackage{booktabs}
\usepackage{multirow}
\usepackage{multicol}
\usepackage{lipsum}
\usepackage{inconsolata}
\usepackage{xcolor}
\usepackage{algorithm}
\usepackage{algorithmic}
\usepackage{amsmath,textcomp,amssymb,geometry,graphicx,enumerate}
\usepackage{cleveref}

\usepackage{array}
\newcolumntype{P}[1]{>{\raggedright\arraybackslash}p{#1}}

\newif\ifcomments
\ifcomments
    \newcommand{\vivek}[1]{{\protect\color{orange}{[VV: #1]}}}
    \newcommand{\evef}[1]{{\protect\color{blue}{[EF: #1]}}}
    \newcommand{\nick}[1]{{\protect\color{red}{[NT: #1]}}}
    \newcommand{\dan}[1]{{\protect\color{pink}{[DK: #1]}}}
    \newcommand{\todo}[1]{{\color{red}{#1}}}
\else
    \newcommand{\vivek}[1]{}
    \newcommand{\evef}[1]{}
    \newcommand{\nick}[1]{}
    \newcommand{\dan}[1]{}
    \newcommand{\todo}[1]{}
\fi

\usepackage{times}
\usepackage{latexsym}

\usepackage[T1]{fontenc}

\usepackage[utf8]{inputenc}

\usepackage{microtype}

\usepackage{inconsolata}

\title{Ghostbuster: Detecting Text Ghostwritten by Large Language Models}

\author{Vivek Verma \qquad Eve Fleisig \qquad Nicholas Tomlin \qquad Dan Klein \\
  Computer Science Division, UC Berkeley \\
  \texttt{\{vivekverma, efleisig, nicholas\_tomlin, klein\}@berkeley.edu}}

\begin{document}
\maketitle
\begin{abstract}
We introduce Ghostbuster, a state-of-the-art system for detecting AI-generated text.
Our method works by passing documents through a series of weaker language models, running a structured search over possible combinations of their features, and then training a classifier on the selected features to predict whether documents are AI-generated.
Crucially, Ghostbuster does not require access to token probabilities from the target model, making it useful for detecting text generated by black-box or unknown models.
In conjunction with our model, we release three new datasets of human- and AI-generated text as detection benchmarks in the domains of student essays, creative writing, and news articles. We compare Ghostbuster to several existing detectors, including DetectGPT and GPTZero, as well as a new RoBERTa baseline. Ghostbuster achieves 99.0~F1 when evaluated across domains, which is 5.9~F1 higher than the best preexisting model. It also outperforms all previous approaches in generalization across writing domains (+7.5~F1), prompting strategies (+2.1~F1), and language models (+4.4~F1). We also analyze our system's robustness to a variety of perturbations and paraphrasing attacks, and evaluate its performance on documents by non-native English speakers.
\end{abstract}

\section{Introduction}

Language models such as ChatGPT are capable of producing a wide range of fluent text that closely approximates human language use. However, the proliferation of these models has raised concerns about the authenticity and trustworthiness of text across a variety of domains. For example, concerns that students are submitting assignments \textit{ghostwritten} by language models have led many schools to adapt by restricting the use of ChatGPT in classrooms and homework assignments \citep{school-chatgpt}. Meanwhile, because language models are prone to factual errors and hallucination, readers may desire to know if such tools have been used to ghostwrite news articles or other informative text when deciding whether to trust a source.

Several detection frameworks have been proposed to address this issue, such as DetectGPT \citep{mitchell2023detectgpt} and GPTZero \citep{gptzero}. While these frameworks offer some level of detection, we found that their performance falters on datasets that they were not originally evaluated on (Section~\ref{sec:results}). In addition, the high false positive rates of these models raise potential ethical concerns because they jeopardize students whose genuine work is misclassified as AI-generated. Furthermore, previous work has indicated that text by non-native speakers of English is disproportionately flagged as AI-generated \citep{LIANG2023100779}. These concerns underscore the need for AI-generated text detectors with strong generalization performance.

\begin{figure*}[h]
\centering
\includegraphics[width=0.95\textwidth]{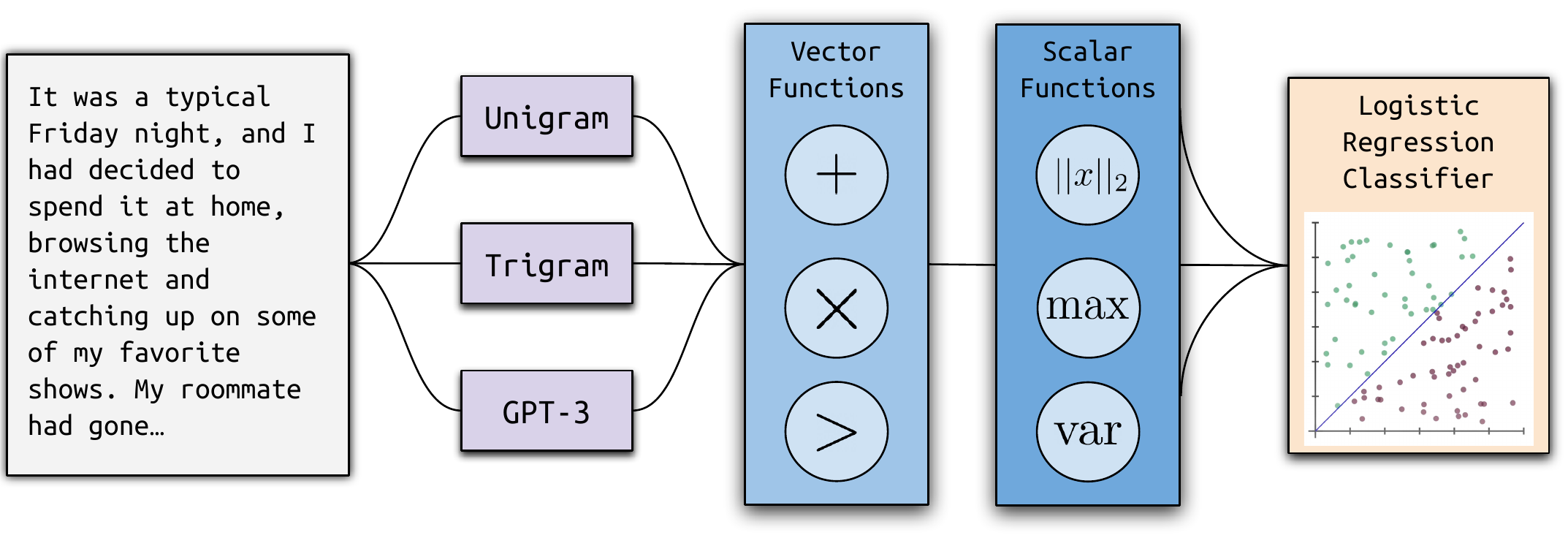}
\caption{An outline of our model training procedure. First, we fed each document into a series of weaker language models to obtain token probabilities. Then, we ran a structured search over combinations of the model outputs and trained a linear classifier on the selected features.}
\label{fig:overview}
\end{figure*}

We present Ghostbuster, a method for detection based on structured search and linear classification (\Cref{fig:overview}). First, Ghostbuster passes paired human-authored and AI-generated documents through a series of weaker language models, ranging from a unigram model to the non-instruction-tuned GPT-3 \texttt{davinci}. Given the word probabilities from these models, it then searches over a space of vector and scalar functions that combine these probabilities into a small set of features. Finally, it feeds these features into a linear classifier, as described in Section~\ref{sec:model}. Our model obtains 99.0~F1 on in-domain classification, outperforming DetectGPT and GPTZero by an average margin of 23.7~F1.
We also evaluated Ghostbuster's generalization to new datasets, prompting strategies, and models  (\Cref{sec:results}) and conducted a number of ablations and robustness experiments (\Cref{sec:analysis}). 
Ghostbuster is available at \url{ghostbuster.app}; we release code for our method and replicating our experiments at \url{github.com/vivek3141/ghostbuster}.

\section{Related Work}

Language models exhibit only some statistical properties of human-authored text. \citet{meister2021language} found that AI-generated text adhered to Heaps' type-token law less than human-authored text, but was similar to human text in terms of unigram distribution, length, stopword usage, and adherence to Zipf's rank-frequency law. \citet{ippolito2020automatic} found that improved decoding strategies, such as top-k sampling, fooled humans more often but  introduced statistical abnormalities that made generated text easier to detect with models. AI-generated text also differs qualitatively, and often subtly, from human-authored text. \citet{jawahar2020automatic} found that GPT-2 responses that a model misclassified as human-authored tended to be short and contained issues of factuality, repetition, contradiction, and incoherence; \citet{dou2022gpt3} found that larger models had fewer  errors in areas such as factuality or coherence. \citet{guo2023close} also found that ChatGPT answers were more formal and focused than human ones.


\begin{figure*}
\centering
\includegraphics[width=0.95\textwidth]{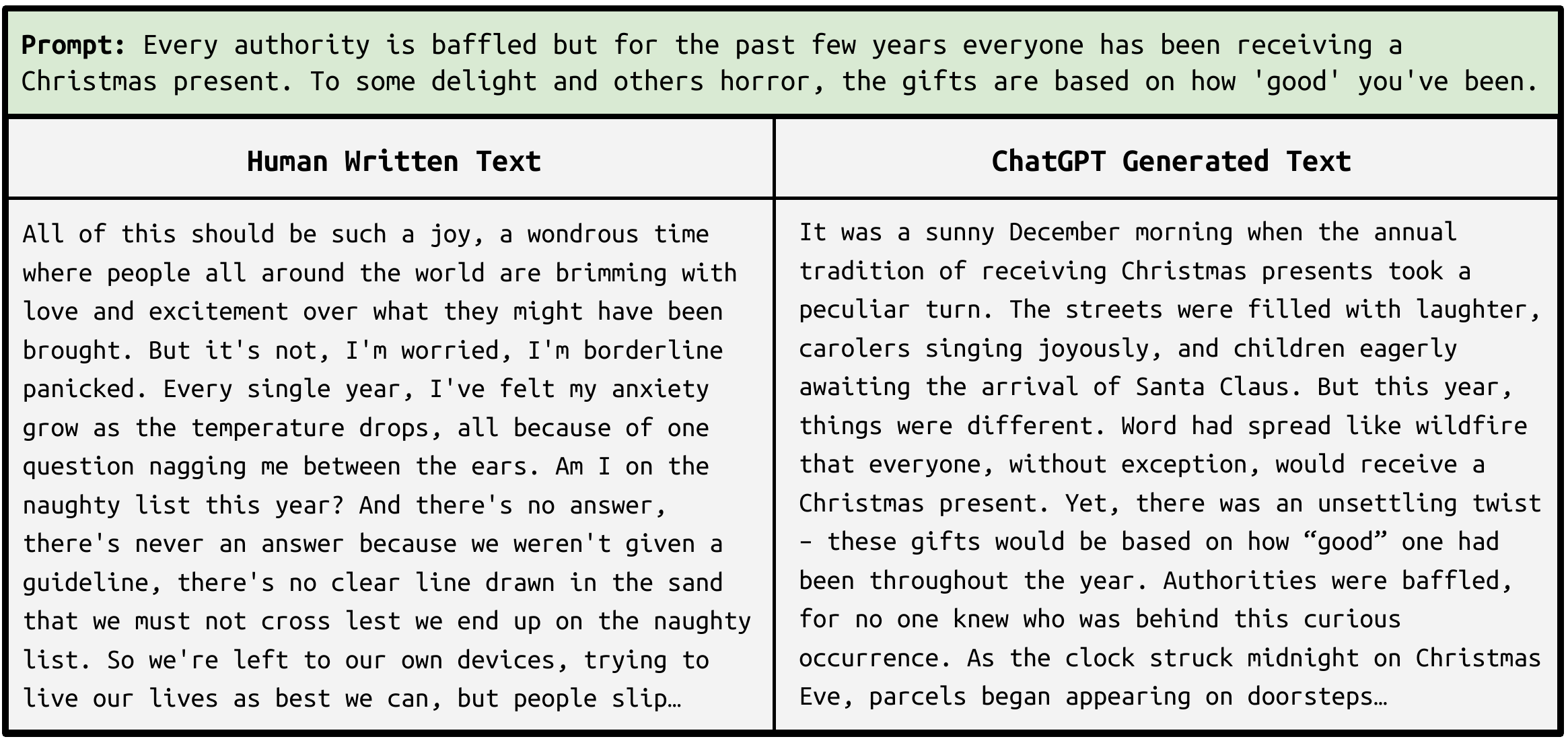}
\caption{An example comparison between human-written and ChatGPT-written text. Whenever possible, all ChatGPT-generated documents were generated based on the same prompts seen by human writers. When prompts were not available, we used ChatGPT to first generate a prompt conditioned on the human-written text and then used ChatGPT again to generate a document conditioned on the generated prompt.}
\end{figure*}

AI-generated text also exhibits qualitative differences from human-authored text, though these are often subtle. \citet{jawahar2020automatic} found that GPT-2 responses that a model misclassified as human authored tended to be very short and contained issues of factuality, repetition, contradiction, and incoherence; however, \citet{dou2022gpt3} had human labelers annotate AI-generated text and found that larger models were less likely to produce responses that were factually incorrect, contradicted common sense, or were incoherent, while the degree of self-contradiction and redundant text varies substantially depending on the model. More recently, \citet{guo2023close} found that while volunteers often rated ChatGPT answers as more helpful than human ones, ChatGPT answers were still more formal, more strictly focused, and used more conjunctions. 

One line of work aims to ensure that AI-generated text can be detected through deliberate watermarking of AI outputs \citep{aaronson2023, pmlr-v202-kirchenbauer23a, zhao2023provable, kamaruddin}. Watermarking has the benefit of providing guarantees on the probability that text is successfully detected, though it relies on devlelopers incorporating watermarks into models.

Several tools have been introduced to detect AI-generated text in the absence of watermarking. \citet{gehrmann2019gltr} introduced GLTR, a suite of statistical tools to aid humans in detecting AI-generated text, which include overlaying text with the text's top-k annotation in different colors. DetectGPT \citep{mitchell2023detectgpt} used the fact that unlike human-authored text, generated text lies in regions of the probability space where nearby samples often have lower model probability. Their model generates random perturbations of the text from a generic LM to detect AI-generated text, then gets probabilities of the original text and perturbations from the model that might have generated the text. Recent supervised methods have used logistic regression, RoBERTa, and T5 to distinguish between human-authored and AI-generated text \citep{guo2023close, chen2023gptsentinel, uchendu-etal-2020-authorship}. Concurrent with this work, \citet{bhattacharjee2023conda} used contrastive domain adaptation for unsupervised AI-generated text detection.

However, \citet{sadasivan2023aigenerated} argued that there is an upper bound on the performance of generated text detectors and found that many are brittle to paraphrasing attacks, including DetectGPT \citep{mitchell2023detectgpt}, GLTR \citep{gehrmann2019gltr}, and other zero-shot methods \citep{ippolito2020automatic, solaiman2019release}, as well as OpenAI’s generated text detectors \citep{openai2019}.
In this paper, we focus primarily on the setting in which entire paragraphs or documents were generated by language models, leaving adversarial prompting and paraphrasing attacks as an avenue for future work. In addition, \citet{LIANG2023100779} found that essays by non-native English speakers were disproportionately misclassified as AI-generated by seven commercial GPT-generated text detectors. In \Cref{sec:ethics}, we discuss Ghostbuster's performance on non-native English speaker data and other ethical considerations regarding detection models.


\section{Datasets}
We collected three new datasets for benchmarking detection of AI-generated text across the domains of creative writing, news, and student essays. For each of the three datasets, we collected ChatGPT-generated text corresponding to the human-authored text. All training datasets were generated using \texttt{gpt-3.5-turbo}.

Our creative writing dataset is based on the subreddit \texttt{r/WritingPrompts}, a forum in which users share creative writing prompts and craft stories in response to these prompts. In order to avoid contamination from ChatGPT-written content, we collected data from the top 50 posters in October 2022 and scraped the last 100 posts by each of these users.\footnote{ChatGPT was publicly released on November 30, 2022.} We used the scraped prompts to generate ChatGPT-written essays.

Our news dataset is based on the Reuters 50-50 authorship identification dataset \citep{Houvardas2006NGramFS}, which consists of 5000 news articles by 50 journalists. Because we did not have access to ground truth headlines or summaries for these articles, we first prompted ChatGPT to generate a headline for each article (see \Cref{sec:prompts}), then prompted it to write an article based on each generated headline.

Finally, our student essay dataset is based on essays from IvyPanda, which consists of high school and university level essays across a range of disciplines. As with the news dataset, we did not have access to the original prompts for these essays, so we used ChatGPT to first generate a prompt corresponding to the essay and then generate a corresponding essay that responds to that prompt. In order to avoid spurious effects of document length, for each of the three domains, we also prompted ChatGPT to approximately match the length of the corresponding human-written article, as shown in \Cref{tab:prompts}.

Additional details on the prompting process and dataset statistics are provided in \Cref{sec:prompts}. For each task, the datasets were divided into train, validation, and test sets. To validate task difficulty and ensure no major artifacts remained in the datasets, we asked human reviewers to label subsets of the essays as human or AI generated (see \Cref{sec:human-eval}).

\paragraph{Evaluation Datasets}
To evaluate generalization across different models and prompting strategies, we collected Claude-generated text based on the original prompt in \Cref{tab:prompts}, as well as ChatGPT-generated text across a variety of additional prompts. Only human and ChatGPT-written essays from the original prompt in \Cref{tab:prompts} were used for model training, and all other data was used solely for evaluation.
Because reducing the false positive rate is particularly important for applications such as detecting student use of AI-generated text, we evaluated accuracy on some datasets of human text alone (i.e., a precision-only evaluation), including several datasets of text by non-native English speakers (details in \Cref{sec:ethics}).

\section{Model}
\label{sec:model}
Given the concerns of brittleness raised for many current AI-generated text detection systems \citep[][\textit{inter alia}]{bakhtin2019real, pu2023deepfake, sadasivan2023aigenerated}, a principal objective for Ghostbuster was to train a model with strong generalization ability across a variety of distribution shifts, including different text domains, prompts, and models. Two of our baselines, a perplexity-only model and a RoBERTa-based model, represent the extremes that Ghostbuster aims to avoid (see baselines in \Cref{sec:baselines}). The simplest approach, using document perplexity alone to distinguish AI-generated and human-authored text, resulted in a brittle and insufficiently expressive model with particularly poor performance on domain and style shifts, including worse-than-random accuracy on some out-of-domain tasks and on some non-native English speaker data (see \Cref{sec:results}). 
However, we also found that highly expressive models such as RoBERTa can exhibit catastrophic worst-case performance, performing even worse than the perplexity-only baseline for certain domain shifts (up to 21.3~F1 drop). We designed  
Ghostbuster to take a middle path, using combinations of features based on the probabilities of documents under a series of language models that are weaker than the target model. This combination of operations provides a comparatively simple way to add model capacity without creating a fully neural architecture, which is more likely to overfit. 

Ghostbuster uses a three-stage training process (see \Cref{fig:overview}): \textit{probability computation}, \textit{feature selection}, and \textit{classifier training}. First, we converted each document into a series of vectors by computing per-token probabilities under a series of language models. Then, we selected features by running a structured search procedure over a space of vector and scalar functions that combine these probabilities. To do so, we defined a set of operations that combine these features and ran forward feature selection on them. Finally, we trained a simple classifier on the best probability-based features and some additional manually-selected features.

\subsection{Probability Computation}
\label{sec:probabilities}
Ghostbuster first passes each document through a series of language models that are weaker than the target model to compute vectors of token probabilities for each document. Our approach uses a unigram fertility model, a Kneser-Ney trigram model, and two early GPT-3 models (\texttt{ada} and \texttt{davinci}, without instruction tuning) to obtain these probabilities. We describe additional details of model training in \Cref{sec:algorithm}.

\begin{table}[t]
    \centering
    \begin{tabular}{llllr}
      \toprule
      \textbf{Vector Functions} & \textbf{Scalar Functions}\\
      \midrule
      $f_{\text{add}_i} = p_{1_i} + p_{2_i}$ & $f_{\text{max}} = \max{p}$ \\
      $f_{\text{sub}_i} = p_{1_i} - p_{2_i}$ & $f_{\text{min}} = \min{p}$ \\
      $f_{\text{mul}_i} = p_{1_i} \cdot p_{2_i} $ & $f_{\text{avg}} = {1 \over |p|} \sum_i p_i$ \\
      $f_{\text{div}_i} = p_{1_i} / p_{2_i}$ & $f_{\text{avg-top25}} = {1 \over |p|} \sum_{i \in T_p} p_i$\\
      $f_{\text{>}_i} = \mathbbm{1}_{\{p_{1_i} > p_{2_i}\}}$ & $f_{\text{len}} = |p|$\\
      $f_{\text{<}_i} = \mathbbm{1}_{\{p_{1_i} < p_{2_i}\}}$ & $f_{\text{L2}} = ||p||_2$\\
      & $f_{\text{var}} = {1 \over n} \sum_{i} (p_i - \mu_p)^2$\\
      \bottomrule
    \end{tabular}
    \caption{List of vector and scalar functions used for feature generation. Vector functions take in two vectors of probabilities $p_1, p_2 \in \mathbb{R}^n$ and output a single vector $f \in \mathbb{R}^n$, where $n$ is the number of tokens in a document. On the other hand, scalar functions take in an input vector $p \in \mathbb{R}^n$ and output $f \in \mathbb{R}$. Here, $T_p$ denotes the indices that contain the top 25 lowest values in $p$ and $\mu_p$ denotes the average value of $p$.}
    \label{tab:func}
\end{table}

\begin{algorithm}[t]
\begin{algorithmic}
\caption{Subroutine \textsc{find-all-features}}
\label{alg:feature}
\REQUIRE The previously picked feature $p$, depth $d \leq \texttt{max\_depth}$, vectors $V$ of token probabilities (from unigram, trigram, \texttt{ada}, and \texttt{davinci} models), scalar functions $F_s$, vector functions $F_v$
\ENSURE A list of all possible features

\STATE Let $S = \emptyset$
\FORALL{scalar functions $f_s \in F_s$}
\STATE{Add $f_s(p)$ to $S$}
\ENDFOR
\FORALL{combinations of features and vector functions $(p^{\prime}, f_v) \in V \times F_v$}
\STATE {Add $\textsc{find-all-features}(f_v(p, p^{\prime}), d+1)$ to $S$}
\ENDFOR
\end{algorithmic}
\end{algorithm}

\subsection{Feature Selection}
\label{sec:feature_selection}

Feature selection proceeded in two stages: we first generated a set of features and then combined them using \Cref{alg:feature}. To generate features, we first constructed 13 vector and scalar operations, which are outlined in \Cref{tab:func}. 
The scalar operations (such as \texttt{avg} or \texttt{var}) convert vectors to scalars, and vector operations (such as \texttt{add}) combine two vectors into one. In order to generate all possible features, we ran Algorithm~\ref{alg:feature} four times, with the probability vectors from each model as the starting features and a maximum depth of $3$. Features thus took the form of combining three arbitrary vectors of probabilities with vector functions, then reducing them to a scalar function. An example feature is \texttt{var(unigram\_probs > ada\_probs - davinci\_probs)}. This approach defines a \textit{structured search space}, in which only a limited set of easily interpretable features are used as input to our classification model. We provide more details on the implementation and outputs of the algorithm in Appendix~\ref{sec:algorithm}. For a version of Ghostbuster trained on each dataset, we ran forward feature selection to find the best features, as listed in Appendix~\ref{sec:best_features}.

\begin{table*}[h]
\small
  \begin{center}
    \begin{tabular}{lp{1.1cm}p{1.1cm}p{1.1cm}p{1.1cm}p{1.1cm}p{1.1cm}p{1.1cm}}
    
    \toprule
       & \multicolumn{4}{c}{\textbf{In-Domain}} & \multicolumn{3}{c}{\textbf{Out-of-Domain}} \\
       \cmidrule(l{5pt}r{5pt}){2-5}
       \cmidrule(l{5pt}r{5pt}){6-8}
       \textbf{Model} & All \break Domains & News & Creative Writing & Student \break Essays & News & Creative Writing & Student \break Essays \\
      \midrule
      Perplexity only & 81.5 & 82.2 & 84.1 & 92.1 & 71.9 & 49.0 & 93.4 \\
      DetectGPT & 57.4 & 56.6 & 48.2 & 67.3 & 56.6 & 48.2 & 67.3 \\
      GPTZero & 93.1 & 91.5 & 93.1 & 83.9 & 91.5 & 93.1 & 83.9 \\
      RoBERTa & 98.1 & 99.4 & 97.6 & 97.4 & 88.3 & \textbf{95.7} & 71.4 \\
      \textbf{Ghostbuster} & \textbf{99.0} & \textbf{99.5} & \textbf{98.4} & \textbf{99.5} & \textbf{97.9} & 95.3 & \textbf{97.7} \\
      \bottomrule
    \end{tabular}
  \end{center}
  \caption{Results of our model across a variety of text domains (F1). We first trained and evaluated Ghostbuster on each of three domains individually (news, creative writing, or student essays); in the ``All Domains'' condition, Ghostbuster was trained and evaluated on all three domains at once. For each out-of-domain setting, Ghostbuster was trained on two domains and evaluated on one held-out domain. Because DetectGPT and GPTZero are unsupervised methods, their performance does not differ across the in-domain and out-of-domain conditions. Out-of-domain performance therefore yields the fairest comparison across models.}
  \label{tab:combined_results}
\end{table*}

\subsection{Classifier Training}
\label{sec:classifier}
Ghostbuster's classifier was trained on combinations of the probability-based features chosen through structured search, as well as seven additional features (\Cref{sec:additional_features}) based on word length and the largest token probabilities. These additional features are intended to incorporate qualitative heuristics observed about AI-generated text. 

The classifier itself is a logistic regression classifier trained with L2 regularization and setting $C=1$ that takes in these features and those chosen through structured search  (\Cref{sec:feature_selection}).

\section{Baselines}
\label{sec:baselines}
We evaluated Ghostbuster's performance relative to multiple existing methods, including unsupervised and supervised detectors, and conducted human evaluation to validate task difficulty.

We compared our model to DetectGPT \citep{mitchell2023detectgpt}, an unsupervised method that generates random perturbations of the text from a generic LM to detect AI-generated text, then gets probabilities of the original text and perturbations from the model that might have generated the text. However, DetectGPT is known to perform poorly when the scoring and target models differ, making it less suitable for detection of text generated from commercial models like ChatGPT or Claude that do not surface token probabilities. Following the implementation in \citet{mitchell2023detectgpt}, we used GPT-2~XL as the scoring model for DetectGPT. 
We also compared with GPTZero \citep{gptzero}, a commercial model that uses a mixture of approaches, including supervised training, perplexity, variance in perplexity, and internet search. 
Because neither DetectGPT nor GPTZero are trained on our datasets, their in-domain and OOD results do not differ, and the generalization experiments provide the fairest comparison to Ghostbuster.
For both DetectGPT and GPTZero, we also performed \textit{oracle thresholding}, where we set the distributions of predicted labels to match those of our test sets for a fairer comparison. However, we did not perform oracle thresholding on the precision-only evaluation in \Cref{sec:ethics}, since it would trivially result in perfect accuracy. We also did not perform oracle thresholding on Ghostbuster or any of the supervised baselines discussed below.

Our simplest supervised baseline is a linear classifier trained only on the perplexities of human-authored and AI-generated documents, as measured by \texttt{davinci}; this classifier learns a single threshold parameter based on its training set. In addition, we fine-tuned a supervised RoBERTa-based model on the same data, similar to the RoBERTa-based approaches in \citet{uchendu-etal-2020-authorship}, \citet{guo2023close}, and \citet{chen2023gptsentinel}. We employed \texttt{roberta-large} with a logistic regression head, and fine-tuned with early stopping. 

\paragraph{Human Evaluation}
\label{sec:human-eval}
We collected human annotations to validate the difficulty of our datasets and provide a human baseline. Six undergraduate and PhD students with previous experience using text generation models were given a random set of 50 documents, evenly sampling human-authored and AI-generated documents, and asked to label whether the documents were written by a human or AI. The average human accuracy was 59\% (maximum = 80\%, minimum = 34\%), suggesting that our task is difficult for humans. 
We then collected additional data via an interface for human evaluation, available at \url{https://ghostbuster.app/experiment}. Through this interface, a total of 233 participants contributed by labeling whether documents were written by a human or AI; of these, 17 participants made at least 25 guesses. Among these 17, the average accuracy (in balanced class binary classification) was 58.1 $\pm$ 11.1\% (maximum = 82.0\%, minimum = 39.7\%).

\begin{table*}[t]
\small
  \begin{center}
    \begin{tabular}{lccccc}
    
    \toprule
      \textbf{Model} & \textbf{Prompts} (F1) & \textbf{Claude} (F1) & \textbf{Lang8} (Acc.) & \textbf{TOEFL 11} (Acc.) & \textbf{TOEFL 91} (Acc.) \\
      \midrule
      Perplexity only & 85.3 & 84.1 & 98.6 & 98.1 & 13.2 \\
      DetectGPT & 70.8 & 64.2 & 98.6 & \textbf{100.0} & 63.7 \\
      GPTZero & 96.1 & 75.6 & \textbf{99.2} & \textbf{100.0} & 92.3 \\
      RoBERTa & 97.4 & 87.8 & 98.6 & 98.1 & \textbf{96.7} \\
      \textbf{Ghostbuster} & \textbf{99.5} & \textbf{92.2} & 95.5 & 99.9 & 74.7 \\
      \bottomrule
    \end{tabular}
  \end{center}
  \caption{Additional generalization results. We evaluated model performance across a variety of prompting strategies (example prompts in \Cref{tab:prompts}). We also evaluated our model's ability to detect essays generated by Claude. Finally, we evaluated the model accuracy on a set of three datasets of text written by non-native English speakers. For all conditions, we trained the RoBERTa and Ghostbuster models on all three domains of human and ChatGPT-written text; we applied oracle thresholding on the DetectGPT and GPTZero models only for the prompt and model generalization experiments. For generalization across models and prompts, we report F1; for the non-native English speaker data, we report accuracy (equivalent to precision, since there is no corresponding AI-generated text).}
  \label{tab:results4}
\end{table*}
    
\begin{table*}[h]
\small
  \begin{center}
    \begin{tabular}{lp{1.1cm}p{1.1cm}p{1.1cm}p{1.1cm}p{1.1cm}p{1.1cm}p{1.1cm}}
    
    \toprule
       & \multicolumn{4}{c}{\textbf{In-Domain}} & \multicolumn{3}{c}{\textbf{Out-of-Domain}} \\
       \cmidrule(l{5pt}r{5pt}){2-5}
      \cmidrule(l{5pt}r{5pt}){6-8}
       \textbf{Ablation} & All\break Domains & News & Creative Writing & Student \break Essays  & News & Creative Writing & Student \break Essays \\
      \midrule
      Handcrafted features only & 80.5 & 79.6 & 78.2 & 83.6 & 75.8 & 77.2 & 77.2 \\
      Limited search (depth $=1$) & 93.7 & 96.9 & 89.6 & 93.9 & 93.7 & 81.3 & 87.3 \\
      Limited search (depth $=2$) & 98.3 & 98.1 & 98.1 & 98.8 & 95.9 & 95.2 & 93.1 \\
      Further search (depth $=4$) & 98.3 & \textbf{99.5} & 97.8 & 99.4 & 96.4 & \textbf{95.8} & \textbf{97.7} \\
      Without \texttt{ada} and \texttt{davinci} & 88.2 & 91.8 & 93.7 & 96.5 & 70.1 & 78.5 & 75.5 \\
      Without \texttt{davinci} & 98.8 & 99.3 & \textbf{99.5} & \textbf{99.8} & 97.3 & 90.3 & 91.9\\
      Without handcrafted features & 98.9 & 99.0 & 98.9 & 99.5 & 97.8 & 93.4 & 97.4 \\
      With random features & 97.8 & 98.7 & 97.3 & 99.4 & 94.3 & 93.1 & 94.3\\
      Ghostbuster (full model) & \textbf{99.0} & \textbf{99.5} & 98.4 & 99.5 & \textbf{97.9} & 95.3 & \textbf{97.7}\\
      \bottomrule
    \end{tabular}
  \end{center}
  \caption{Model ablations (F1). We first evaluated the performance of Ghostbuster using only handcrafted features, without the structured search procedure defined in \Cref{sec:feature_selection}. We also experimented with only allowing one or two operations during structured search, effectively limiting the space of potential features. Finally, we evaluated the performance of our model without access to \texttt{ada} and/or \texttt{davinci}, or without access to any of the handcrafted features, finding that a model which uses only n-gram and \texttt{ada} features (i.e., the without \texttt{davinci} condition) nearly matched the performance of our full model.}
  \label{tab:results3}
\end{table*}

\section{Results}
\label{sec:results}

\subsection{In-domain Classification}
We first evaluated Ghostbuster in-domain, where we trained and classified on the same domain (Table~\ref{tab:combined_results}, left). We found that Ghostbuster achieves 99.0 F1 across all three datasets, outperforming GPTZero by a margin of 5.9~F1 and DetectGPT by 41.6~F1. DetectGPT's weak performance was not unexpected, however, as \citet{mitchell2023detectgpt} reported that accuracy degrades significantly when the scoring and target models differ.
While our RoBERTa baseline achieved an impressive 98.1~F1 when evaluated in-domain on all datasets, it performed inconsistently in our generalization experiments, as discussed in the following sections.

\subsection{Generalization Across Domains}
\label{sec:domaingeneralization}
We also found that Ghostbuster's performance gains over previous models are robust with respect to the similarity of the training and testing datasets.
While Ghostbuster outperformed previous approaches when evaluating and training on the same domain, we note that this comparison is potentially unfair since these datasets are out-of-domain for GPTZero and DetectGPT. \Cref{tab:combined_results} (right) provides results on evaluating Ghostbuster out-of-domain (evaluated on one domain, trained on all other domains). When evaluated out-of-domain, Ghostbuster achieved 97.0~F1 averaged across all conditions, outperforming DetectGPT by 39.6 F1 and GPTZero by 7.5 F1. Ghostbuster outperformed the RoBERTa baseline on all domains except creative writing out-of-domain, and RoBERTa had much worse out-of-domain performance on average (13.8 F1 margin). We also evaluated on non-native English data; performance dropped on one dataset, though this may be largely due to shorter document length (see Section \ref{sec:ethics}).

\subsection{Generalization Across Prompts}
We also evaluated Ghostbuster's robustness to changes in prompts, finding that its performance is maintained when different prompts are used to generate training and evaluation data (Table~\ref{tab:results4}). For each domain, we wrote five different prompts intended to capture natural variation in language model prompting strategies across users, as well as specific stylistic requests users might make to avoid detection, such as asking a model to write in the style of a high-school student.
Ghostbuster achieved 99.5~F1 across prompt variants, compared to 97.4~F1 for RoBERTa and 96.1~F1 for GPTZero. These results suggest that Ghostbuster's performance is not hindered by stylistic or semantic shifts induced by the variations in prompting strategies.

\subsection{Generalization Across Models}
In addition to providing out-of-domain results when generalizing across domains and prompting strategies, we also evaluated Ghostbuster's ability to generalize to different target models. Table~\ref{tab:results4} provides results when evaluating on a Claude-generated dataset. Although Ghostbuster outperformed all other tested approaches with 92.2~F1, the 6.8~F1 decrease in Ghostbuster's performance on Claude data suggests that generalization to different models without training on data from them remains a more challenging task.

\begin{figure*}[h]
\centering
\includegraphics[width=\textwidth]{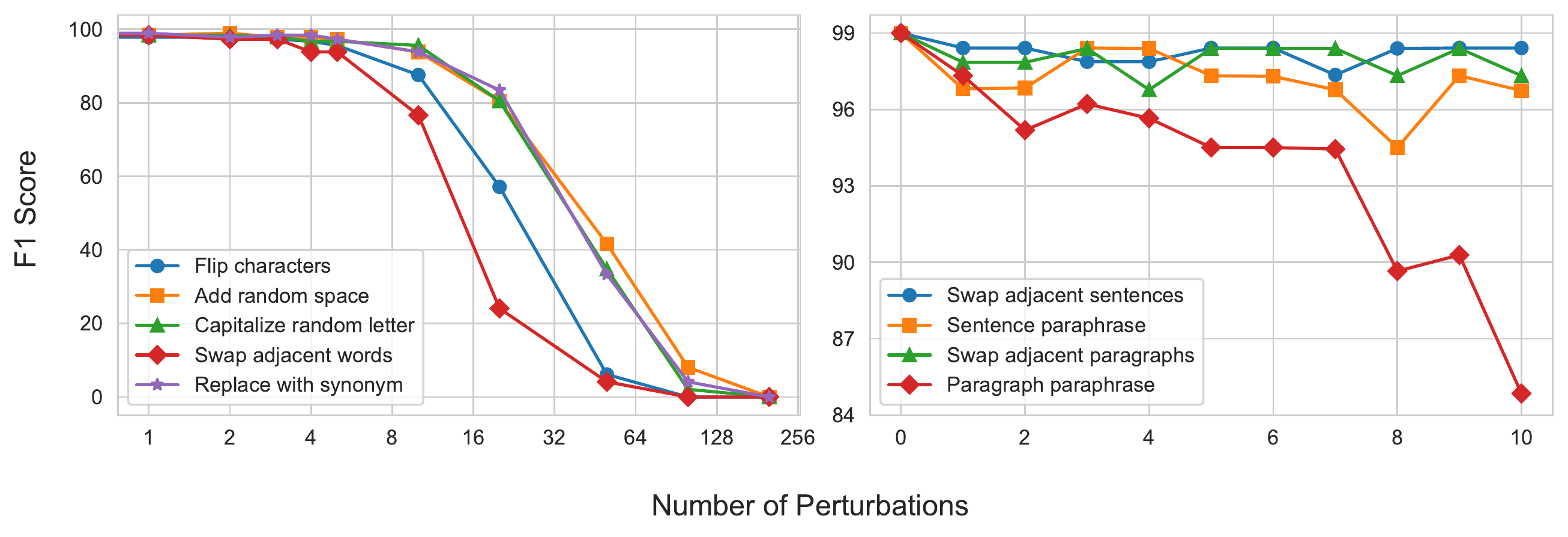}
\caption{Robustness experiments on Ghostbuster (F1). We evaluated the performance of our system on documents that underwent a number of character- and word-level perturbations (left) as well as sentence- and paragraph-level perturbations (right). We describe the details of these perturbations in \Cref{sec:robustness}.}
\label{fig:robustness}
\end{figure*}

\begin{figure}[h]
\centering
\includegraphics[width=\linewidth]{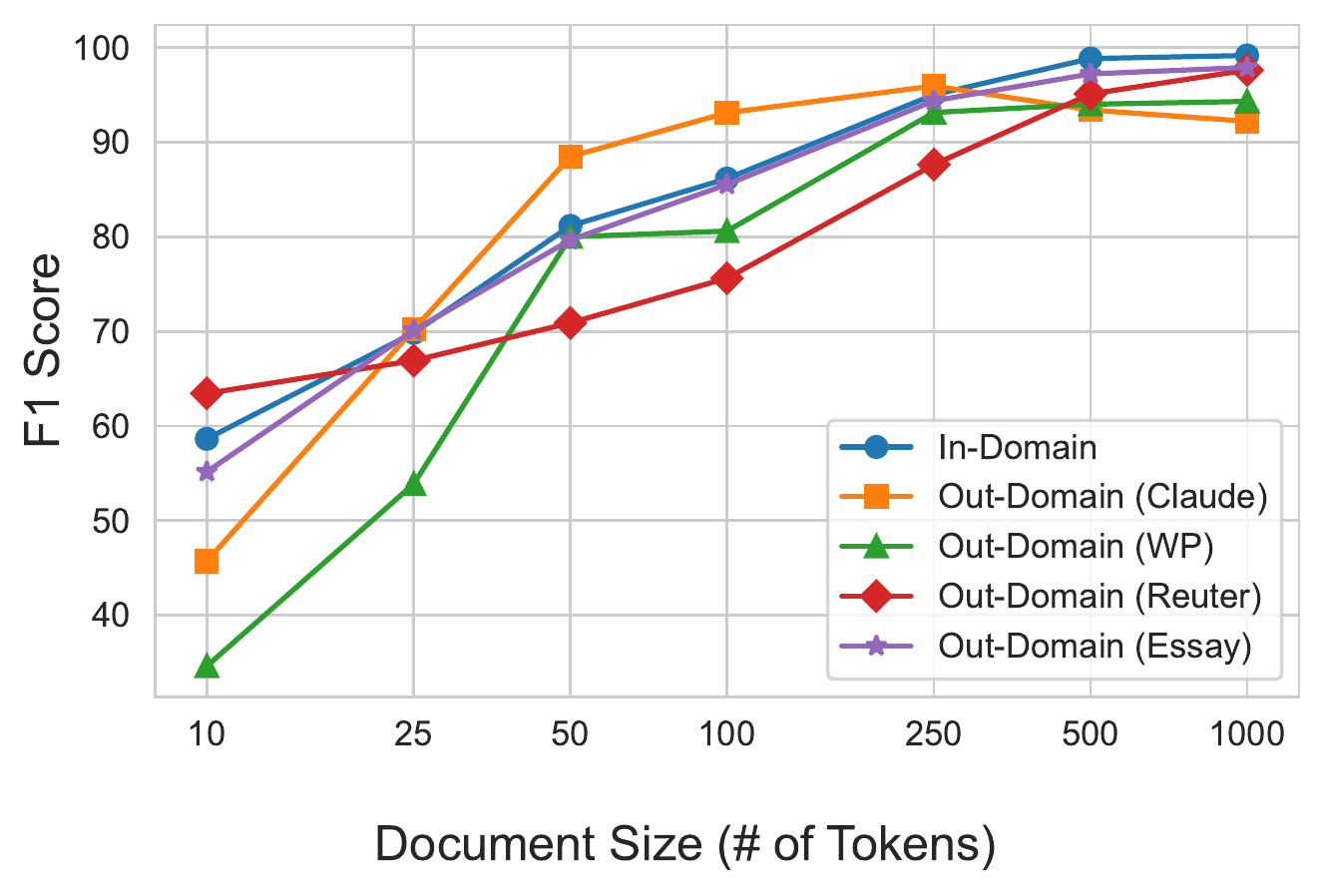}
\caption{Model performance across document lengths, for various domain shifts (F1). We evaluated our model on trimmed documents. Our model predictably performed better on longer documents, whether evaluated in-domain or out-of-domain.
We note that Ghostbuster may be unreliable for documents with $\le 100$ tokens, and its performance levels off with $\ge 500$ tokens.}
\label{fig:length}
\end{figure}

\section{Analysis}
\label{sec:analysis}

To understand the performance of Ghostbuster in more depth, we performed ablations that suggest the structured search and use of probabilities from a neural LLM are crucial to performance (\Cref{sec:ablations}). We also ran experiments on robustness to editing essays under different perturbations, finding that most global edits do not significantly affect performance and numerous local edits are required to fool the model. Analyzing performance on shorter documents, we found that Ghostbuster's  performance improved with document length, and was generally more reliable on documents with more than 100 tokens.

\subsection{Ablations}
\label{sec:ablations}
We conducted multiple ablations to understand the role of the depth of structured search over features, probabilities from different models, and type of features used (structured search or handcrafted) on model performance. We observed that conducting structured search with depth lower than 3 tends to underfit the data, whereas performance plateaus or degrades for search depths greater than 3 (Table~\ref{tab:results3}). In addition, we saw that omitting the usage of \texttt{davinci} results in out-of-domain generalization performance decreasing by 0.6 to 5.8~F1. This suggests that incorporating token probabilities from a model closer in quality to the target model improves generalization. Training the model without \texttt{ada} and \texttt{davinci} probabilities (i.e., using unigram and trigram features only) resulted in performance decreases of up to 10.8~F1 in-domain and 27.8~F1 out-of-domain, highlighting the importance of probabilities from a neural LLM for model performance, and particularly generalization performance. We also observed that while in-domain performance remains similar without handcrafted features, generalization performance decreases by 0.1 to 2.9~F1, suggesting that the handcrafted features play a more limited role in preventing overfitting. By contrast, removing the structured search features and training on only the handcrafted features decreased performance by 15.9 to 22.1~F1, suggesting that the structured search is crucial.

\subsection{Robustness}
\label{sec:robustness}
We also ran a series of robustness experiments in which we evaluated the performance of Ghostbuster on modified essays, under a number of different perturbation functions. These experiments were intended to determine whether our detector can be easily evaded by simply misspelling a word or adding nonsense tokens at the beginning of the document.
In \Cref{fig:robustness}, we plot both local perturbations (character and word-level; left) and global perturbations (sentence or paragraph level; right). Our local perturbations consisted of randomly (1) adding a character, deleting a character, or swapping a pair of adjacent characters; (2) adding or removing space or newline characters; (3) capitalizing or lowercasing letters; (4) swapping adjacent words; and (5) replacing words with their closest synonyms, using NLTK's \texttt{wordnet.synset} implementation. Our global perturbations consisted of (6) swapping adjacent sentences; (7) swapping adjacent paragraphs; (8) replacing sentences with paraphrases; and (9) replacing paragraphs with their paraphrases.\footnote{We used \url{https://huggingface.co/tuner007/pegasus_paraphrase} as the paraphrase model.} In general, we found that performance smoothly degrades when performing local edits, and that numerous edits are typically required to produce a false negative. Most global edits did not significantly affect model performance, e.g., swapping adjacent sentences or paragraphs had a negligible effect. However, repeated calls to a paraphrasing model will lead to worse results. We also experimented with the commercial detection evader Undetectable AI (see \Cref{sec:algorithm}), finding that  recall degraded from 99\% to 62\%. While targeted attacks like these may lead to more misclassifications than random perturbations on the input, we find that sites like Undetectable AI still fool our model less than half of the time. Full details of this experiment are available in \Cref{sec:algorithm}. 

\subsection{Non-Native English Speaker Data}
\citet{LIANG2023100779} sampled 91 TOEFL\footnote{Test of English as a Foreign Language, an exam taken by non-native speakers of English to attend English-speaking universities.} essays and found that more than half of the essays were misclassified as AI-generated by seven GPT-generated text detectors. We evaluated Ghostbuster's performance on three datasets of non-native English speaker data: the TOEFL 11 dataset \citep{Blanchard2013ETS}, the 91 TOEFL essays that \cite{LIANG2023100779} tested on, and the Lang8 dataset \citep{mizumoto-etal-2011-mining-lang8} (details in \Cref{sec:non-native-data}). 

We found that all tested models had over 95\% accuracy, but experienced significant performance drops on the 91 TOEFL essays from previous work. This may be partly due to the fact that the 91 TOEFL essays are significantly shorter (median 104 words) than the TOEFL 11 essays (median 315 words) and our other datasets (all above median 350 words). Ghostbuster also experienced some performance drop on the Lang8 dataset, which also consists of shorter documents (median Lang8 document length is 77 words). This performance is similar to Ghostbuster's out-of-domain performance on documents of similar length: performance on out-of-domain documents with 100 words ranged from 75.6 to 93.1 F1, compared to 74.7 F1 for the 91 TOEFL essays. This suggests that the performance drop on the 91 TOEFL essays and on Lang8 can be largely, though not entirely, attributed to the challenge of classifying shorter documents.

\subsection{Performance Across Document Lengths}
Previous work has shown that AI-generated text detectors perform better on longer documents \citep[e.g.,][]{solaiman2019release, chakraborty2023possibilities}. We replicated this finding by training Ghostbuster on full-length documents and running it on documents trimmed to $N$ tokens, for each value of $N \in [10, 25, 50, 100, 250, 500, 1000]$. We found that document length had a similar effect on both in-domain and out-of-domain performance (\Cref{fig:length}). In line with prior work, we found model performance degraded substantially on documents with $\le 100$ tokens, making it difficult to determine whether individual paragraphs of documents were generated by language models. To encourage future work on this issue, we propose an additional paragraph-level detection benchmark in \Cref{sec:addbenchmarks}.

\section{Conclusion}
We introduced Ghostbuster, a model for detecting AI-generated language that uses structured search on token probabilities from weaker models to identify whether a given document was AI-generated. We validated Ghostbuster by evaluating its performance on datasets from three domains (news, student essays, and creative writing), as well as through generalization experiments on text generated by different models and using different prompts. We also release our three datasets as benchmarks for evaluating performance on detecting AI-generated text. Ghostbuster achieved over 98.4~F1 across all datasets on in-domain detection of AI-generated text, representing substantial progress over currently available models.

Future work could examine tradeoffs between the false positive and false negative rates of AI-generated text detectors for different applications. For detection of AI-generated student essays, lowering the risk of false positives is a key priority to avoid false accusations of student misconduct. In other settings, however, false positives are less concerning. For example, if detectors are used to prevent AI-generated text from being used in language model training data, or to flag potentially AI-generated content on the web, the ideal model calibration may differ. Other avenues for future work include improving robustness to perturbations of AI-generated outputs, such as lightly editing to avoid detection, and different task formulations, including detection at the paragraph level for documents that combine human-authored and AI-generated text. Finally, future AI-generated text detectors could provide additional explanations for classification decisions, so that human users can use their own judgment when evaluating the decisions of these systems.

\newpage
\section{Ethics and Limitations}
\label{sec:ethics}

We trained and evaluated Ghostbuster on three datasets that represent a range of domains, but note that these datasets are not representative of all writing styles or topics and contain predominantly British and American English text. Thus, incorrect predictions by Ghostbuster may be particularly likely for text that represents a distributional shift from Ghostbuster's training. Issues relating to improving model performance on shorter text, a broader range of domains, varieties of English besides American and British English, and robustness to edits are important areas for future work. 

More broadly, users wishing to apply Ghostbuster to real-world cases of potential off-limits usage of text generation (e.g., identifying ChatGPT-written student essays) should be aware that incorrect predictions by Ghostbuster may be particularly likely for shorter text, domains further from those on which Ghostbuster was trained (e.g., text messages), text in varieties of English besides Standard American or British English, or in non-English languages, text written by non-native speakers of English, AI-generated text that has been edited or paraphrased by a human, and text that was generated by prompting an AI model to paraphrase or adjust a human-authored input. To avoid perpetuation of algorithmic harms due to these limitations, we strongly discourage incorporation of Ghostbuster into any systems that automatically penalize students or other writers for alleged usage of text generation without human supervision. Instead, we recommend cautious use of Ghostbuster, in conjunction with human supervision and additional factors, if classifying a person's writing as AI-generated could harm that person. Ghostbuster can also be used for a variety of lower-risk applications, including filtering AI-generated text out of language model training data and checking whether potential online sources of information are AI-generated.

\section*{Acknowledgments}
This work was supported by grants from Open Philanthropy, DARPA SemaFor, and the Bakar Spark program. We are especially grateful to Lucy Li for her help with data collection. We also thank Naitian Zhou, Kyle Mahowald, Robin Netzorg, and the members of the Berkeley NLP Group for establishing human baselines, stress-testing our system, and providing writing feedback and suggestions. 

\bibliography{references}

\newpage
\appendix

\section{Prompting and Dataset Details}
\label{sec:prompts}
For the news and creative writing datasets, we first prompted the model to generate a headline or writing prompt, respectively, before generating the documents themselves from those prompts. \Cref{tab:prompt-details} gives the full prompting strategy for the original prompts. \Cref{tab:gen-prompts} gives all the generalization prompts used by dataset. \Cref{tab:datasets} gives details on training dataset sizes and splits, and \Cref{tab:esl-datasets} provides details on additional evaluation datasets used to evaluate the performance of our model on documents written by non-native English speakers.

\section{Non-Native English Speaker Data Descriptions}
\label{sec:non-native-data}

The TOEFL 11 dataset contains university-level essays written on the TOEFL exam, with an even number of essays by authors whose first languages (L1s) were Arabic, Chinese, French, German, Hindi, Italian, Japanese, Korean, Spanish, Telugu, and Turkish. \cite{LIANG2023100779}'s 91 TOEFL essays were drawn from a Chinese educational forum. The Lang8 dataset contains data from an online forum used by language learners from a range of countries, and particularly from Japan. We evaluated Ghostbuster's out-of-the-box performance on 1,000 examples from each of the Lang8 and TOEFL 11 datasets, as well as the 91 TOEFL essays from \cite{LIANG2023100779}.


\begin{table*}[h]
    \centering
    \begin{tabular}{lP{4cm}cP{2cm}P{2cm}P{2cm}}
        \toprule
        \textbf{Domain} & \textbf{Human Text Source} & \# \textbf{Docs} & \multicolumn{3}{c}{\textbf{Median Words per Document}} \\
        \cmidrule(l{5pt}r{5pt}){4-6}
        & & & Human \break (1,000 docs) & ChatGPT \break (5,000 docs) & Claude \break (1,000 docs)\\
        \midrule
        Student Essays & IvyPanda \citep{IvyPanda} & 7,000 & 529 & 559 & 442 \\
        News Articles & Reuters 50-50  \citep{Houvardas2006NGramFS} & 7,000 & 498 & 510 & 384\\
        Creative Writing & r/WritingPrompts & 7,000 & 455 & 512 & 384\\
        \bottomrule
    \end{tabular}

    \caption{Datasets introduced in this paper. For each domain, the 5,000 ChatGPT-generated documents are divided into 1,000 documents from the same prompt, and a 4,000-document ``generalization set'' that used different prompts to evaluate generalization. For each domain, 1,000 human-authored documents and the 1,000 ChatGPT-generated documents that used the same prompt were split into train, validation, and test sets used by Ghostbuster. The ChatGPT-generated generalization set and the Claude-generated articles were used only for evaluation of Ghostbuster, not training.}
    \label{tab:datasets}
\end{table*}

\begin{table*}[h]
    \centering
    \begin{tabular}{lcc}
        \toprule
        \textbf{Source} & \# \textbf{Docs} & \textbf{Median~Words per Document} \\
        \midrule
        Lang8 \citep{mizumoto-etal-2011-mining-lang8} & 1,000 & 77 \\
        TOEFL 11  \citep{Blanchard2013ETS} & 1,000 & 315 \\
        TOEFL from \cite{LIANG2023100779} & 91 & 104 \\
        \bottomrule
    \end{tabular}
    \caption{Non-native English datasets evaluated on in this paper. For these datasets, only the original human data was evaluated on (no parallel data was generated).}
    \label{tab:esl-datasets}
\end{table*}

\begin{table*}[h]
    \centering
    \begin{tabular}{lP{2.9cm}P{8cm}}%
        \toprule
        \textbf{Dataset} & \textbf{Prompting Strategy} & \textbf{Prompt}\\
        \midrule
        Creative writing & Generate story & \texttt{Write a story in \{length\} words to the prompt: \{prompt\}}\\
        \midrule
        \multirow{2}{*}{News Articles} & (1) Generate title  & \texttt{Create a headline for the following news article: \{doc\}} \\
        & (2) Generate article & \texttt{Write a news article in \{length\} words with the following headline \{headline\}.}\\
        \midrule
        \multirow{2}{*}{Student Essay} & 
        (1) Generate prompt &
        \texttt{Given the following essay, write a prompt for it: \{doc\}} \\
        & (2) Generate essay &
        \texttt{Write an essay in \{length\} words to the prompt: \{prompt\}}\\
        \bottomrule
    \end{tabular}
    \caption{Prompts used to generate documents in each of the three proposed datasets. Ground truth prompts are available for the \texttt{r/WritingPrompts} data but not for the news or student essay domains, so we use ChatGPT to first generate a corresponding prompt or headline based on the original human-written article and then generate an entire article based solely on the generated prompt in these domains.}
    \label{tab:prompt-details}
\end{table*}

\begin{table*}[h]
    \centering

    \begin{tabular}{P{2.3cm}P{3.7cm}P{3.7cm}P{3.7cm}}
    \toprule
    \textbf{Dataset} & \textbf{Student Essays} & \textbf{News Articles} & \textbf{Creative Writing}\\
    \midrule
     Original Prompt & \texttt{Write an essay in \{length\} words to the prompt: \{prompt\}} & 
      \texttt{Write a news article in \{length\} words with the following headline \{headline\}.} & \texttt{Write a story in \{length\} words to the prompt: \{prompt\}} \\
      Generalization Prompt 1 & \texttt{You are a student, who is writing an essay in response to the prompt \{prompt\}. What would you write in \{length\} words?} & \texttt{You are a news reporter, who is writing an article with the headline \{headline\}. What would you write in \{length\} words?} & \texttt{You are an author, who is writing a story in response to the prompt \{prompt\}. What would you write in \{length\} words?} \\
     Generalization Prompt 2 & \texttt{Hi! I'm trying to write a \{length\}-word essay based on the following prompt: \{prompt\}. Could you please draft something for me?} & \texttt{Hi! I'm trying to write a \{length\}-word news article based on the following headline: \{headline\}. Could you please draft something for me?} & \texttt{Hi! I'm trying to write a \{length\}-word story on the following prompt: \{prompt\}. Could you please draft something for me?} \\
     Generalization Prompt 3 & \texttt{Write a \{length\}-word essay in the style of a high-school student  in response to the following prompt: \{prompt\}.} & \texttt{Write a \{length\}-word news article in the style of a New York Times article based on the headline \{headline\}.} & \texttt{Write a \{length\}-word story in the style of a beginner writer in response to the prompt \{prompt\}.} \\
     Generalization Prompt 4 & \texttt{Write an essay with very short sentences in \{length\} words to the prompt \{prompt\}.} & \texttt{Write a news article with very short sentences in \{length\} words based on the headline \{headline\}.} & \texttt{Write a story with very short sentences in \{length\} words to the prompt \{prompt\}.} \\
     \bottomrule
\end{tabular}

    \caption{Full set of prompts used to produce paired ChatGPT-generated data. We set \texttt{length} equal to the number of words in the human-authored example rounded to the nearest 100, and prompt with the \texttt{prompt} or \texttt{headline} corresponding to each story. This approach is intended to prevent document length or content effects from trivializing the detection task. In the student essay and news domains, we follow the procedure described in \Cref{tab:prompt-details} in \Cref{sec:prompts} to obtain the prompts or headlines used to generate articles.}
    \label{tab:gen-prompts}
    \label{tab:prompts}
\end{table*}

\section{Additional Implementation Details}
\label{sec:algorithm}

\paragraph{Feature Selection Algorithm} In Algorithm~\ref{alg:feature}, we describe the algorithm used to compute all possible features. We perform a brute-force approach, pruning out features which apply the same function twice. For each depth $d$, we consider all possible combination of $d-1$ vector functions, then append all scalar functions on top.



\paragraph{Unigram and Trigram Models} In order to obtain vectors of identical length, the unigram and trigram models are both trained over the GPT-3 tokenizer vocabulary, without \texttt{UNK}ing; the Kneser-Ney model uses a discount factor of $\delta = 0.9$ for backoff to both the bigram and unigram fertility model. The unigram and trigram models were both trained using counts from the Brown Corpus \citep{francis1979brown}.

\paragraph{Pruning Candidate Features}
While the algorithm presented in Algorithm~\ref{alg:feature} produces equivalent results to our implementation, we make a simple additional optimization. Because many vector functions in Table~\ref{tab:func} are commutative, we prune our list of possible vector combinations to avoid double-counting. This pruning results in around a 2/3rd reduction in the feature space. At a search depth of 3, we have 2534 available features to select from; at a search depth of 2, we have 322 available features.

\paragraph{Undetectable.ai} We ran additional robustness experiments using the commercial detection evader service \url{https://undetectable.ai}. We randomly selected 100 AI-generated essays across each of the three domains (news, fiction, and student essays) and fed them into the Undetectable API. For documents from the writing prompts dataset, we use the \texttt{university} readability setting and the \texttt{story} purpose; for documents from the Reuters dataset, we use the \texttt{journalist} readability setting and the \texttt{article} purpose; and for the student essay dataset, we use the \texttt{high school} readability setting and the \texttt{essay} purpose. We use the \texttt{balanced} setting across all three datasets, which controls the extent of text modification.


\section{Additional Features}
\label{sec:additional_features}
Ghostbuster uses the following handcrafted features in addition to those chosen through feature selection:
\begin{itemize}
    \item Number of outliers ($p_i > 10$), average value of top 25 and 25-50$^{\text{th}}$ largest token probabilities
    \item Average value of the 25 largest and 25-50$^{\text{th}}$ largest token probabilities of the vector $d - a$, where $d$ is a vector of \texttt{davinci} token probabilities and $a$ is a vector of \texttt{ada} token probabilities.
    \item Average length of the 25 longest and 25-50$^{\text{th}}$ longest words, measured in tokens.
\end{itemize}

\section{Best Features}
\label{sec:best_features}
\begin{figure*}[h]
\centering
\includegraphics[width=\textwidth]{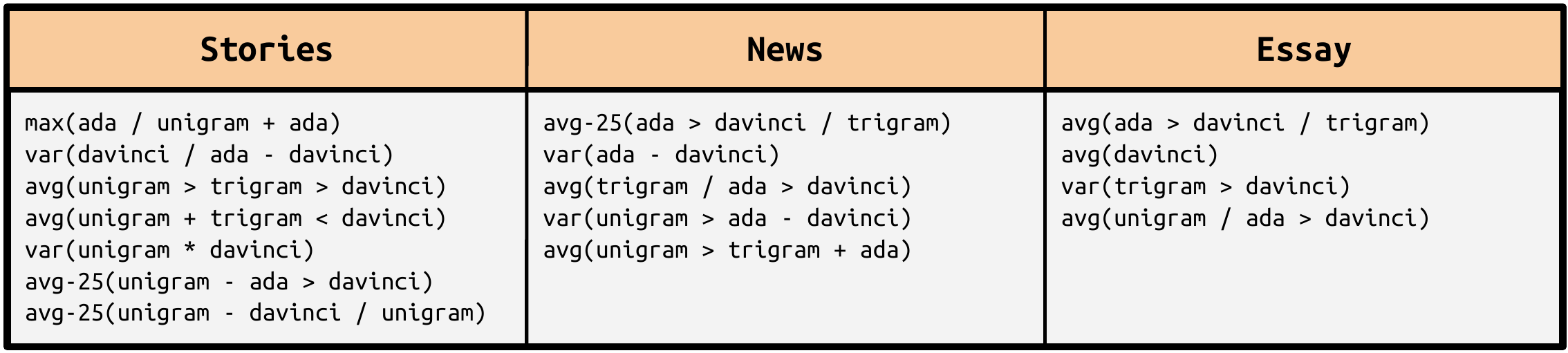}
\caption{A list of features selected when performing forward feature selection on each of the three domains. 
We show in \Cref{tab:results3} that these features lead to significantly better performance than using randomly chosen features or classifying based on our set of manually-selected features alone.
}
\label{fig:features}
\end{figure*}

In Figure~\ref{fig:features}, we present the best features chosen through validation on each of the datasets. For a list of functions and features used, refer to Table~\ref{tab:func}.

\section{Additional Benchmarks}
\label{sec:addbenchmarks}
Whether detection of AI-generated text is a trivial or intractable task depends greatly on the framing of the problem. Generated text that contains human paraphrases or was generated from particularly clever prompts is especially difficult to detect with machine learning methods \citep{sadasivan2023aigenerated}. However, trained humans have consistently been able to spot ChatGPT's style \citep{guo2023close}.

We introduce benchmarks for several framings of the detection task, with ascending levels of difficulty: \textit{author identification},  detecting whether a document was AI-generated or written by a single author, given a history of documents by that author; \textit{document-level detection}, detecting whether a full document was AI-generated; and \textit{paragraph-level detection}, detecting which paragraphs in a document were AI-generated.

These tasks are motivated by real-world applications of these detectors. For example, when questioning whether a student assignment was AI-generated, instructors often have access to previous work by that student; document-level detection may be useful when instructors do not have access to a history of student writing, or cannot verify that students did not include AI-generated text in previous assignments; and paragraph-level detection is useful when, as is often the case, AI assistance was only used for portions of the assignment. Although the modeling work in the main body of our paper focuses only on document-level detection, we will release these additional benchmarks as a testbed for future work at \url{https://github.com/vivek3141/ghostbuster-data/}

\begin{figure*}
\centering
\includegraphics[width=\textwidth]{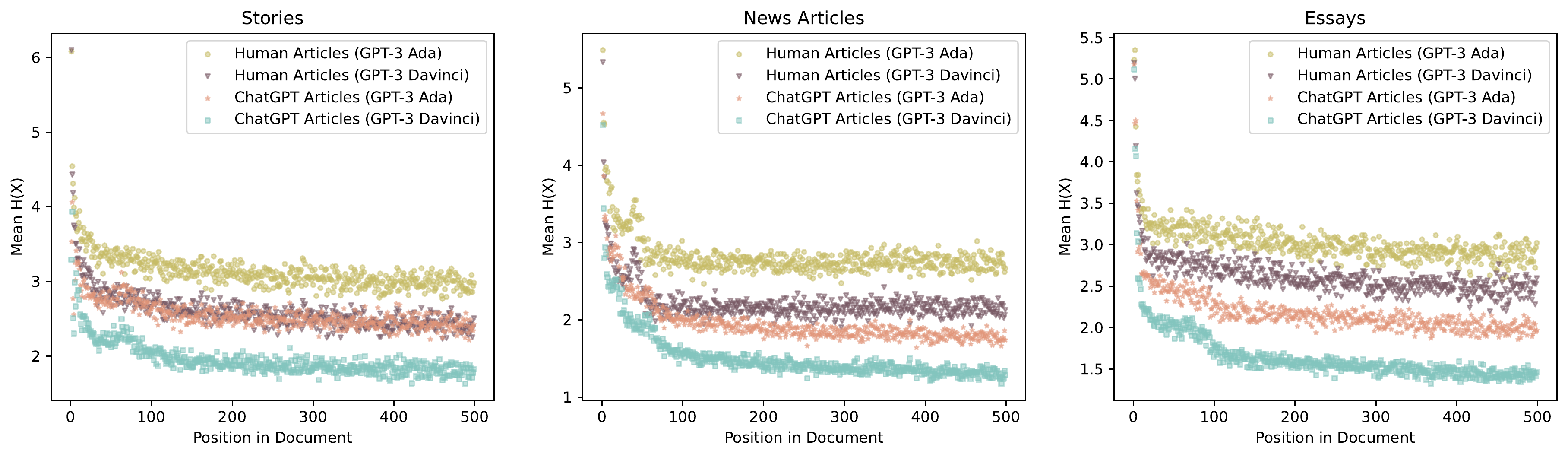}
\caption{Entropy rate of Writing Prompts, Reuters and IvyPanda under GPT-3 (Ada and Davinci). We notice that entropy values sharply decline towards the beginning, with a plateau or slight decline towards the end. We also notice that ChatGPT documents tend to be more predictable than human documents, but this difference is more pronounced towards the end of documents.}
\label{fig:entropy}
\end{figure*}

\section{Qualitative Analysis of Trends in Token Probabilities}

In this section, we visualize a subset of the vectors fed into our model, motivated by \citet{verma-etal-2023-revisiting}. For each document in a given domain, we plot the average per-token probability across documents in the dataset. For instance, at token position $i$, we compute the value
\[
f(i) = {1 \over |W|} \cdot \sum_{w \in W} \log P_{\theta}(w_i)
\]
where $w$ denotes a document in an article $W$, $w_i$ denotes the $i^{\text{th}}$ token in that document and $P_{\theta}(w_i) = P_{\theta}(w_i \ | \ w_{i-1}, ..., w_1)$ is the conditional probability given by GPT-3. We refer to the trend of $f(i)$ as \textit{entropy rate}, although we note that this definition differs from the classic definition of entropy rate in information theory.

As shown in \Cref{fig:entropy}, both human documents and ChatGPT documents follow a similar trend to the one described in \citet{verma-etal-2023-revisiting}, with a sharp decline towards the beginning of documents, followed by a gradual decline or a plateau towards the end of the document. Furthermore, we observe that documents written by ChatGPT tend to be more predictable than human written documents, but the gap between human and ChatGPT documents increases towards the end. We believe that this trend provides an example of the type of distributional difference that Ghostbuster may be able to leverage during classification.
\end{document}